\documentclass[runningheads]{llncs}
\usepackage{graphicx}
\usepackage{comment}
\usepackage{amsmath,amssymb} 
\usepackage{color}

\usepackage{bm}
\usepackage{mathrsfs}

\begin{document}
\pagestyle{headings}
\mainmatter
\def\ECCVSubNumber{1742}  

\title{LEED: Label-Free Expression Editing via Disentanglement} 

\titlerunning{LEED: Label-Free Expression Editing via Disentanglement}
%
\author{Rongliang Wu \and
Shijian Lu}

\authorrunning{Rongliang Wu and Shijian Lu}

\institute{Nanyang Technological University \\
\email{ronglian001@e.ntu.edu.sg, shijian.lu@ntu.edu.sg}
} 

\maketitle

\begin{abstract}
Recent studies on facial expression editing have obtained very promising progress. On the other hand, existing methods face the constraint of requiring a large amount of expression labels which are often expensive and time-consuming to collect.
This paper presents an innovative label-free expression editing via disentanglement (LEED) framework that is capable of editing the expression of both frontal and profile facial images without requiring any expression label. The idea is to disentangle the identity and expression of a facial image in the expression manifold, where the neutral face captures the identity attribute and the displacement between the neutral image and the expressive image captures the expression attribute. Two novel losses are designed for optimal expression disentanglement and consistent synthesis, including a mutual expression information loss that aims to extract pure expression-related features and a siamese loss that aims to enhance the expression similarity between the synthesized image and the reference image. 
Extensive experiments over two public facial expression datasets show that LEED achieves superior facial expression editing qualitatively and quantitatively.

\end{abstract}

\section{Introduction}

Facial expression editing (FEE) allows users to edit the expression of a face image to a desired one.
Compared with facial attribute editing which only considers appearance modification of specific facial regions~\cite{zhang2018generative,li2016deep,shen2017learning}, FEE is much more challenging as it often involves large geometrical changes and requires to modify multiple facial components simultaneously. FEE has attracted increasing interest due to the recent popularity of digital and social media and a wide spectrum of applications in face animations, human-computer interactions, etc.

Until very recently, this problem was mainly addressed from a graphical perspective in which a 3D Morphable Model (3DMM) was first fitted to the image and then re-rendered with a different expression~\cite{geng20193d}. Such methods typically involve tracking and optimization to fit a source video into a restrictive set of facial poses and expression parametric space~\cite{wu2018reenactgan}. A desired facial expression can be generated by combining the graphical primitives~\cite{li2010example}.
Unfortunately, 3DMMs can hardly capture all subtle movements of face with the pre-defined parametric model and often produce blurry outputs due to the Gaussian assumption~\cite{geng20193d}.

Inspired by the recent success of Generative Adversarial Nets (GANs)~\cite{goodfellow2014generative}, a number of networks~\cite{qiao2018geometry,song2018geometry,ding2018exprgan,pumarola2018ganimation,choi2018stargan,geng2018warp,geng20193d,wu2020cascade} have been developed and achieved very impressive FEE effects. Most of these networks require a large amount of training images with different types of expression labels/annotations, e.g. discrete labels~\cite{choi2018stargan,ding2018exprgan}, action units intensity~\cite{pumarola2018ganimation,wang2019dft,wu2020cascade} and facial landmarks~\cite{qiao2018geometry,song2018geometry,geng2018warp}, whereas labelling a large amount of facial expression images is often expensive and time-consuming which has impeded the advance of relevant research on this task.
At the other end, the ongoing research ~\cite{pumarola2018ganimation,qiao2018geometry,song2018geometry,geng20193d,wu2020cascade} is largely constrained on the expression editing of frontal faces due to the constraint of existing annotations, which limits the applicability of FEE in many tasks. 

This paper presents a novel label-free expression editing via disentanglement (LEED) framework that can edit both frontal and profile expressions without requiring any expression label or annotation by humans. Inspired by the manifold analysis of facial expressions~\cite{chang2006manifold,ekman2002facial,jiang2019disentangled} that different persons have analogous expression manifolds, we design an innovative disentanglement network that is capable of separating the identity and expression of facial images of different poses. 
The label-free expression editing is thus accomplished by fusing the identity of an input image with an arbitrary expression and the expression of a reference image. Two novel losses are designed for optimal identity-expression disentanglement and identity-preserving expression editing in training the proposed method. The first loss is a mutual expression information loss that guides the network to extract pure expression-related features from the reference image. The second loss is a siamese loss that enhances the expression similarity between the synthesized image and the reference image. Extensive experiments show that our proposed LEED even outperforms supervised expression editing networks qualitatively and quantitatively.

The contributions of this work are threefold. First, we propose a novel label-free expression editing via disentanglement (LEED) framework that is capable of editing expressions of frontal and profile facial images without requiring any expression label and annotation by humans. 
Second, we design a mutual expression information loss and a siamese loss that help extract pure expression-related features and enhance the expression similarity between the edited and reference facial images effectively. Third, extensive experiments show that the proposed LEED is capable of generating high-fidelity facial expression images and even outperforms many supervised networks.

\begin{figure}[t]
\begin{center}
\includegraphics[width=1.\linewidth]{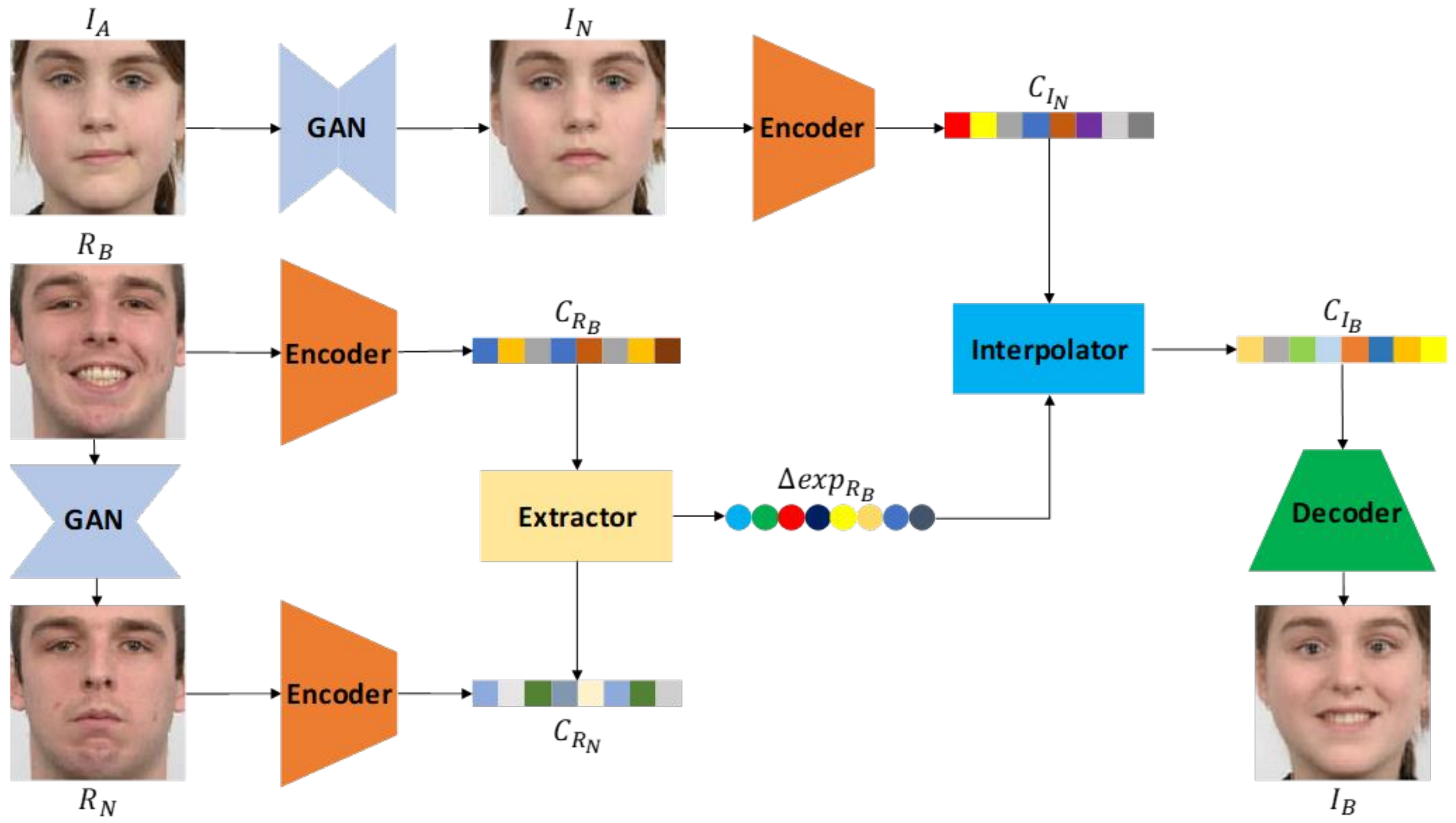}
\end{center}
\caption{
The framework of LEED: given an input image $I_A$ and a reference image $R_B$, the corresponding neutral faces $I_N$ and $R_N$ are first derived by a pre-trained GAN. The \textit{Encoder} maps $I_N$, $R_N$, and $R_B$ to latent codes $C_{I_N}$, $C_{R_N}$, and $C_{R_B}$ in an embedded space which capture the identity attribute of $I_A$, the identity attribute of $R_B$, and the identity and expression attributes of $R_B$, respectively. The \textit{Extractor} then extracts the expression attribute of $R_B$ ($\Delta {exp}_{R_B}$) from $C_{R_B}$ and $C_{R_N}$, and the \textit{Interpolator} further generates the target latent code $C_{I_B}$ from $\Delta {exp}_{R_B}$ and $C_{I_N}$. Finally, the \textit{Decoder} projects $C_{I_B}$ to the image space to generate the edited image $I_B$.
} 
\label{fig:overall}
\end{figure}

\section{Related Work}

\noindent{\bf Facial Expression Editing:} FEE is a challenging task and existing works can be broadly grouped into two categories. The first category is more conventional which exploits graphic models for expression editing. A typical approach is to first fit a 3D Morphable Model to a face image and then re-render it with a different expression. A pioneering work of Blanz and Vetter~\cite{blanz1999morphable} presents the first public 3D Morphable Model. Vlasic et al.~\cite{vlasic2006face} proposes a video based multilinear model to edit the facial expressions. 
Cao et al.~\cite{cao2013facewarehouse} introduces a video-to-image facial retargeting application that requires user interaction for accurate editing. Thies et al.~\cite{thies2016face2face} presents Face2Face for video-to-video facial expression retargeting which assumes the target video contains sufficient visible expression variation.

The second category exploits deep generative networks~\cite{goodfellow2014generative,kingma2013auto}. 
For example, 
warp-guided GAN~\cite{geng2018warp} and paGAN~\cite{nagano2018pagan} are presented to edit the expression of frontal face images with neutral expression. 
G2-GAN~\cite{song2018geometry} and GCGAN~\cite{qiao2018geometry} adopt facial landmarks as geometrical priors to control the generated expressions, where ground-truth images are essential for extracting the geometrical information. \cite{geng20193d} proposes a model that combines 3DMM and GAN to synthesize expressions on RGB-D images.
ExprGAN~\cite{ding2018exprgan} introduces an expression controller to control the intensity of the generated expressions conditioned on the discrete expression labels. StarGAN~\cite{choi2018stargan} generates new expression through identity-preservative image-to-image translation and it can only generate discrete expressions. GANimation~\cite{pumarola2018ganimation} adopts Action Units~\cite{friesen1978facial} as expression labels and can generate expressions in continuous domain. 
Cascade EF-GAN~\cite{wu2020cascade} also uses Action Units and introduces local focuses and progressive editing strategy to suppress the editing artifacts.  
Recently, \cite{qian2019make} proposes AF-VAE for face animation where expressions and poses can be edited simultaneously according to the landmarks boundary maps extracted offline with other tools.

The works using graphics models for expression editing require 3D face scans and/or video sequences as well as efforts and dedicated designs for complex parametric fitting. Additionally, they cannot model invisible parts in the source image such as teeth of a closed mouth. The work using generative networks are more flexible but they require a large amount of labelled expressive images to train the models. Besides, the existing deep generative networks also require suitable expression labels/annotations for guiding the model to synthesize desired expression, where the annotations are either created by humans or extracted from reference images by offline tools. The proposed LEED employs generative networks which can hallucinate missing parts of input face images and it just requires a single photo for the input image and reference image which makes it much simpler to implement. At the same time, it can edit the expression of both frontal facial images and profile images without requiring any expression label/annotation by either humans or other tools.

\noindent{\bf Disentangled Representations:} The key of learning disentangled representation is to model the distinct, informative factors of variations in the data~\cite{bengio2013representation}. Such representations have been applied successfully to image editing~\cite{jiang2019disentangled,narayanaswamy2017learning,shu2017neural}, image-to-image translation~\cite{yang2019disentangling} and recognition tasks~\cite{wang2018orthogonal,peng2017reconstruction,wu2019disentangled}. However, previous works achieve disentangled learning by training a multi-task learning model~\cite{jiang2019disentangled,yang2019disentangling,wang2018orthogonal}, where labels for each disentangled factors are essential. Recently, the unsupervised setting has been explored~\cite{chen2016infogan,higgins2017beta}. InfoGAN~\cite{chen2016infogan} achieves disentanglement by maximizing the mutual information between latent variables and data variation and $\beta$-VAE~\cite{higgins2017beta} learns the independent data generative factors by introducing an adjustable hyper-parameter $\beta$ to the original VAE objective function. But these methods suffer from the lack of interpretability, and the meaning of each learned factor is uncontrollable.
Based on the expression manifold analysis~\cite{chang2006manifold}, our proposed method seeks another way to disentangle the identity and expression attributes from the facial images.

\section{Proposed Method}

\subsection{Overview}
Our idea of label-free expression editing via disentanglement (LEED) is inspired by the manifold analysis of facial expressions~\cite{chang2006manifold,ekman2002facial,jiang2019disentangled} that the expression manifold of different individuals is analogous. On the expression manifold, similar expressions are points in the local neighborhood with a `neutral' face as the central reference point. Each individual has its neutral expression that corresponds to the original point in its own expression manifold and represents the identity attribute. The displacement of an expressive face and its neutral face gives the expression attribute.

Our proposed method achieves label-free expression editing by learning to disentangle the identity and expression attributes and fusing the identity of the input image and the expression of the reference image for synthesizing the desired expression images. As illustrated in Fig.~\ref{fig:overall}, our network has five major components: an extractor for extracting expression attribute; an interpolator for fusing the extracted expression attribute and the identity attribute of the input image; an encoder for mapping the facial images into a compact expression and identity embedded space; a decoder for projecting the interpolated code to image space and a pre-trained GAN for synthesizing the neutral faces.

\subsection{Extractor and Interpolator}
\label{Extractor_Interpolator}

\noindent {\bf{Learning Expression Attribute Extractor:}} Given an input image with arbitrary expression A (denoted as $I_A$) and a reference image with desired expression B (denoted as $R_B$), our goal is to synthesize a new image $I_B$ that combines the identity attribute of $I_A$ and expression attribute of $R_B$. Without the expression labels, our proposed method needs to address two key challenges: 1) how to extract identity attribute from the input image and expression attribute from the reference image, and 2) how to combine the extracted identity and expression attributes properly to synthesize the desired expression images. We address the two challenges by learning an expression attribute extractor \bm{$\mathcal{X}$} and an interpolator \bm{$\mathcal{I}$}, more details to be shared in the following texts.

The label-free expression editing is achieved by disentangling the identity and expression attributes. Given $I_A$ and $R_B$, LEED first employs a pre-trained GAN to generate their corresponding neutral faces $I_N$ and $R_N$. An encoder \bm{$E$} is then employed to map all the images to a latent space, producing $C_{I_A}$, $C_{I_N}$, $C_{R_B}$ and $C_{R_N}$, where $C_{I_A}$/$C_{R_B}$ and $C_{I_N}$/$C_{R_N}$ are the latent codes of the input/reference image and its neutral face, respectively. More details of the pre-trained GAN and \bm{$E$} are to be discussed in Sec.~\ref{fundamentals}.

According to~\cite{chang2006manifold}, the latent code of the neutral face ($C_{I_N}$) represents the identity attribute of the input image, and the displacement between $C_{R_B}$ and $C_{R_N}$ represents the expression attribute of the reference image:
\begin{equation}
    \begin{aligned}
        \Delta {exp}_{R_B}^* = C_{R_B} - C_{R_N}.
        \label{formula:expression_attribute}
    \end{aligned}
\end{equation}
On the other hand, $\Delta {exp}_{R_B}^*$ depends on the embedded space, and the residual between $C_{R_B}$ and $C_{R_N}$ may contain expression-unrelated information such as head-poses variations that could lead to undesired changes in the synthesized images. We therefore propose to learn the expression attribute with an extractor \bm{$\mathcal{X}$} rather than directly using $\Delta {exp}_{R_B}^*$.

Formally, we train an expression extractor \bm{$\mathcal{X}$} to extract the expression $\Delta {exp}_{R_B}$ from $C_{R_{B}}$ and $C_{R_{N}}$ with $\Delta {exp}_{R_B}^*$ as the pseudo label:
\begin{equation}
    \begin{aligned}
        \min_{\bm{\mathcal{X}}} \mathcal{L}_{exp} = 
                \| \Delta {exp}_{R_B} - \Delta {exp}_{R_B}^* \|^2,
        \label{formula:exp}
    \end{aligned}
\end{equation}
where $\Delta {exp}_{R_B} = \bm{\mathcal{X}} (C_{R_B}, C_{R_N})$.

In addition, we design a mutual expression information loss to encourage the extractor to extract pure expression-related information. Specifically, we first use a pre-trained facial expression classification model $\Psi$, i.e. the ResNet~\cite{he2016deep} pre-trained on Real-world Affective Faces Database~\cite{li2017reliable}, 
to extract the features from $R_B$. As such a model is trained for classification task, the features of the last layers contain rich expression-related information~\cite{upchurch2017deep}. We take the features from penultimate layer as the representation of the expression attribute of $R_B$ and denote it as $F_{R_B}$, where $F_{R_B} = \Psi(R_B)$. As $\Psi$ is used for extracting the features, we do not update its parameters in the training process.

In information theory, the mutual information between $A$ and $B$ measures the reduction of uncertainty in $A$ when $B$ is observed. If $A$ and $B$ are related by a deterministic, invertible function, the maximal mutual information is attained~\cite{chen2016infogan}. By maximizing the mutual information between $\Delta {exp}_{R_B}$ and $F_{R_B}$, the extractor will be encouraged to extract pure expression-related features and ignore expression-unrelated information. However, directly maximizing the mutual information is hard as it requires access to the posterior distribution. We follow~\cite{chen2016infogan,ding2018exprgan} to impose a regularizer \bm{$Q$} on top of the extractor to approximate it by maximizing its derived lower bound~\cite{barber2003algorithm}:
\begin{equation}
    \begin{aligned}
        \min_{\bm{Q, \mathcal{X}}} \mathcal{L}_{\bm{Q}} = 
              - {\mathbb{E}} [ {\rm log} (\bm{Q} ( \Delta {exp}_{R_B} | F_{R_B})],
        \label{formula:Q}
    \end{aligned}
\end{equation}

By combining Eqs.~(\ref{formula:exp}) and~(\ref{formula:Q}), the overall objective function of \bm{$\mathcal{X}$} is
\begin{equation}
    \begin{aligned}
        \mathcal{L}_{\mathcal{X}} = 
              \mathcal{L}_{exp} + \lambda_{Q}\mathcal{L}_{\bm{Q}},
        \label{formula:extractor}
    \end{aligned}
\end{equation}
where $\lambda_{Q}$ is the hyper-parameter to balance the terms.

\noindent {\bf {Learning Interpolator:}}
With the identity attribute $C_{I_N}$ of the input image and the expression attribute $\Delta {exp}_{R_B}$ of the reference image, we can easily obtain the latent code $C_{I_B}$ for the target image through linear interpolation
\begin{equation}
    \begin{aligned}
        C_{I_B}^* = C_{I_N} + \Delta {exp}_{R_B}.
        \label{formula:exp_add}
    \end{aligned}
\end{equation}
On the other hand, the linearly interpolated latent code may not reside on the manifold of real facial images and lead to weird editing (e.g. ghost faces) while projected back to the image space. Hence, we train an interpolator \bm{$\mathcal{I}$} to generate interpolated codes and impose an adversarial regularization term on it (details of the regularization term to be discussed in Sec.~\ref{Encoder_Decoder}) as follows:
\begin{equation}
    \begin{aligned}
        \min_{\bm{\mathcal{I}}} \mathcal{L}_{interp} = 
                  \mathcal{L}_{{adv}_{\bm{E,\mathcal{I}}}} + \| \bm{\mathcal{I}} (C_{I_N}, \alpha \Delta {exp}_{R_B}) 
                   - (C_{I_N} + \alpha \Delta {exp}_{R_B}) \|^2,
        \label{formula:interpolation}
    \end{aligned}
\end{equation}
where $\alpha \in [0,1]$ is the interpolated factor that controls the expression intensity of the synthesized image. We can obtain a smooth transition sequences of different expressions by simply changing the value of $\alpha$ once the model is trained.

In addition, the interpolator should be able to recover the original latent code of the input image given his/her identity attribute and the corresponding expression attribute. The loss term can be formulated as follows:
\begin{equation}
    \begin{aligned}
        \min_{\bm{\mathcal{I}}} \mathcal{L}_{idt} = 
                  {\|} \bm{\mathcal{I}} (C_{I_N}, \Delta {exp}_{I_A}) 
                  - C_{I_A} {\|}^2,
        \label{formula:identitical}
    \end{aligned}
\end{equation}
where $\Delta {exp}_{I_A} = \bm{\mathcal{X}} (C_{I_A}, C_{I_N})$.

The final objective function for the interpolator $\bm{\mathcal{I}}$ is
\begin{equation}
    \begin{aligned}
        \mathcal{L}_{\bm{\mathcal{I}}} = 
              \mathcal{L}_{interp} + \mathcal{L}_{idt}.
        \label{formula:interpolator}
    \end{aligned}
\end{equation}

\subsection{Expression Similarity Enhancement}
\label{siamese}

To further enhance the expression similarity between the synthesized image $I_B$ and the reference image $R_B$, we introduce a siamese network to encourage the synthesized images to share similar semantics with the reference image. The idea of siamese network is first introduced in natural language processing applications~\cite{goldberg2014word2vec} that learns a space where the vector that transforms the word \textit{man} to the word \textit{woman} is similar to the vector that transforms \textit{hero} to \textit{heroine}~\cite{amodio2019travelgan}. In our problem, we define the difference between an expression face and its corresponding neutral face as the expression transform vector. And we minimize the difference between the expression transform vector of $R_B$ and $R_N$ and that of $I_B$ and $I_N$. The intuition is that the transformation that turns a similar expressive face into neutral face should be analogous for different identities, which is aligned with the analysis of expression manifold~\cite{chang2006manifold}. 

Specifically, given reference image with expression B ($R_B$), its corresponding neutral face ($R_N$), synthesized image with expression B ($I_B$) and its corresponding neutral face ($I_N$), we first map them into a latent space by the siamese network \bm{$S$} and obtain the transform vectors:
\begin{equation}
    \begin{aligned}
        v_R = \bm{S}(R_B) - \bm{S}(R_N),
    \end{aligned}
\end{equation}
\begin{equation}
    \begin{aligned}
        v_I = \bm{S}(I_B) - \bm{S}(I_N),
    \end{aligned}
\end{equation}
then we minimize the difference between $v_R$ and $v_I$:
\begin{equation}
    \begin{aligned}
      \min_{\bm{S}} \mathcal{L}_{\bm{S}}  = Dist(v_R, v_I),
      \label{formula:siamese}
    \end{aligned}
\end{equation}
where $Dist$ is a distance metric. We adopt cosine similarity as the distance measurement and incorporate the siamese loss in learning the encoder.

\subsection{Encoder, Decoder and GAN}
\label{fundamentals}

\noindent {\bf{Learning Encoder and Decoder:}}
\label{Encoder_Decoder}
Given a collection of facial images $I$, we train an encoder to map them to a compact expression and identity embedded space to facilitate the disentanglement. We aim to obtain a flattened latent space so as to generate smooth transition sequences of different expressions by changing the interpolated factor (Sec.~\ref{Extractor_Interpolator}). This is achieved by minimizing the Wasserstein distance between the latent codes of real samples and the interpolated ones.

Specifically, a discriminator \bm{$\mathcal{D}$} is learned to distinguish the real samples and the interpolated ones and the encoder \bm{$E$} and interpolator \bm{$\mathcal{I}$} are trained to fool the discriminator. We adopt the WGAN-GP~\cite{gulrajani2017improved} to learn the parameters. The adversarial loss functions are formulated as
\begin{equation}
    \begin{aligned}
        \min_{\bm{\mathcal{D}}} \mathcal{L}_{{adv}_{\bm{\mathcal{D}}}} = 
                 & 
                    {\mathbb{E}}_{\hat{C} \sim P_{\hat{I}}} [{\rm log}\bm{\mathcal{D}}(\hat{C})]
                    - {\mathbb{E}}_{C \sim P_{data}} [{\rm log}\bm{\mathcal{D}}(C)]  \\
                 & + \lambda_{gp} {\mathbb{E}}_{\tilde{C} \sim P_{\tilde{C}}} 
                    [(\| \nabla_{\tilde{C}}\bm{\mathcal{D}}(\tilde{C}) \|_2 - 1)^2],
        \label{formula:D}
    \end{aligned}
\end{equation}
\begin{equation}
    \begin{aligned}
        \min_{\bm{E,\mathcal{I}}} \mathcal{L}_{{adv}_{\bm{E,\mathcal{I}}}} = 
                    - {\mathbb{E}}_{\hat{C} \sim P_{\hat{I}}} [{\rm log}\bm{\mathcal{D}}(\hat{C})],
        \label{formula:E_I}
    \end{aligned}
\end{equation}
where $C=\bm{E}(I)$ stands for the code generated by the encoder, $\hat{C}$ the interpolated code generated by the interpolator \bm{$\mathcal{I}$}, $P_{data}$ the data distribution of the codes of real images, $P_{\hat{I}}$ the distribution of the interpolated ones and $P_{\tilde{C}}$ the random interpolation distribution introduced in~\cite{gulrajani2017improved}.

The model may suffer from `mode collapse' problem if we simply optimize the parameters with Eqs.~(\ref{formula:D}) and~(\ref{formula:E_I}). The encoder learns to map all images to a small latent space where the real and interpolated codes are closed that yields a small Wasserstein distance. To an extreme, the Wasserstein distance could be 0 if the encoder maps all images to a single point~\cite{chen2019homomorphic}. To avoid this trivial solution, we train a decoder \bm{$D$} to project the latent codes back to the image space. We follow~\cite{chen2019homomorphic,li2017universal} to train the decoder with perceptual loss~\cite{johnson2016perceptual} as Eq.~(\ref{formula:Decoder}), and impose an reconstruction constraint on the encoder as Eq.~(\ref{formula:Reconstruction}).
\begin{equation}
    \begin{aligned}
        \min_{\bm{D}} \mathcal{L}_{\bm{D}} = 
                    {\mathbb{E}} ({\|} \Phi(\bm{D}(C)) - \Phi(I) {\|}^2),
        \label{formula:Decoder}
    \end{aligned}
\end{equation}
\begin{equation}
    \begin{aligned}
        \min_{\bm{E}} \mathcal{L}_{recon} = 
                    {\mathbb{E}} ({\|} \Phi(\bm{D}(\bm{E}(I))) - \Phi(I) {\|}^2),
        \label{formula:Reconstruction}
    \end{aligned}
\end{equation}
where $\Phi$ is the VGG network~\cite{simonyan2014very} pre-trained on ImageNet~\cite{deng2009imagenet}. 

The final objective function of the encoder can thus be derived as follows:
\begin{equation}
    \begin{aligned}
        \mathcal{L}_{\bm{E}} = 
                    \mathcal{L}_{{GAN}_{\bm{E,\mathcal{I}}}}
                    + \lambda_{recon} \mathcal{L}_{recon}
                    + \lambda_{S} \mathcal{L}_{\bm{S}},
        \label{formula:Encoder}
    \end{aligned}
\end{equation}
where $\lambda_{recon}$ and $\lambda_{S}$ are the hyper-parameters. $E$ and $S$ are updated in an alternative manner. 

\noindent {\bf{Pre-training GAN:}}
\label{GAN}
We generate the neutral face of the input and reference images by using a pre-trained GAN which can be adapted from many existing image-to-image translation models~\cite{zhu2017unpaired,choi2018stargan,xiao2018elegant,kim2017learning}. In our experiment, we adopt the StarGAN~\cite{choi2018stargan} and follow the training strategy in~\cite{choi2018stargan} to train the model. The parameters are fixed once the GAN is trained.

\section{Experiments}
\label{experiment}

\subsection{Dataset and Evaluation Metrics}
\label{Dataset}
Our experiments are conducted on two public datasets including Radboud Faces Database (RaFD)~\cite{langner2010presentation} and Compound Facial Expressions of Emotions Database (CFEED)~\cite{du2014compound}.
RaFD consists of 8,040 facial expression images collected from 67 participants. 
CFEED~\cite{du2014compound} contains 5,060 compound expression images collected from 230 participants. 
We randomly sample 90$\%$ images for training and the rest for testing. All the images are center cropped and resized to 128 $\times$ 128.

We evaluate and compare the quality of the synthesized facial expression images with different metrics, namely, Fr\'{e}chet Inception Distance (FID)~\cite{heusel2017gans}, structural similarity (SSIM) index~\cite{wang2004image}, expression classification accuracy and the Amazon Mechanical Turk (AMT) user study results.
The FID scores are calculated between the final average pooling features of a pre-trained inception model~\cite{szegedy2017inception} of the real faces and the synthesized faces, and the SSIM is computed over synthesized expressions and corresponding expressions of the same identity.

\begin{table}[t]
    \caption{
            Quantitative comparison with state-of-the-art methods on datasets RaFD and CFEED by using FID (lower is better) and SSIM (higher is better).
            }

    \begin{center}
    \setlength{\tabcolsep}{4mm}{
        \begin{tabular}{|c|c|c|c|c|c|}
          \hline
          {}             & \multicolumn{2}{|c|}{RaFD} & \multicolumn{2}{|c|}{CFEED} \\
          \hline
          {}             & FID$\downarrow$        & SSIM$\uparrow$          & FID$\downarrow$         & SSIM$\uparrow$ \\
          \hline
          StarGAN~\cite{choi2018stargan}        & 62.51      & 0.8563       & 42.39       & 0.8011 \\
          \cline{1-5}
          GANimation~\cite{pumarola2018ganimation}     & 45.55      & 0.8686       & 29.07       & 0.8088  \\
          \cline{1-5}
          Ours           & \bf{38.20} & \bf{0.8833}  & \bf{23.60}  & \bf{0.8194}\\
          \hline
        \end{tabular}
    }
    \end{center}
    
    \label{table:FID_SSIM}
\end{table}

\begin{table}[t]
    \caption{
            Quantitative comparison with state-of-the-art methods on datasets RaFD and CFEED by using facial expression classification accuracy (higher is better).
            }

    \begin{center}
    \setlength{\tabcolsep}{4mm}{
        \begin{tabular}{|c|c|c|c|c|}
          \hline
          Dataset & Method & R & G & R + G \\
          \hline
          {}    & StarGAN~\cite{choi2018stargan}    & {}    & 82.37      & 88.48 \\
          \cline{2-2} \cline{4-5}
          RaFD  & GANimation~\cite{pumarola2018ganimation} & 92.21 & 84.36      & 92.31 \\
          \cline{2-2} \cline{4-5}
          {}    & Ours       & {}    & \bf{88.67} & \bf{93.25} \\
          \hline
          {}    & StarGAN~\cite{choi2018stargan}    & {}    & 77.80      & 81.87 \\
          \cline{2-2} \cline{4-5}
          CFEED & GANimation~\cite{pumarola2018ganimation} & 88.23 & 79.46      & 84.42 \\
          \cline{2-2} \cline{4-5}
          {}    & Ours       & {}    & \bf{84.35} & \bf{90.06} \\
          \hline
        \end{tabular}
    }
    \end{center}
    
    \label{table:cls_accuracy}
\end{table}

\begin{table}[t]
    \caption{
            Quantitative comparison with state-of-the-art methods on RaFD and CFEED by AMT based user studies (higher is better for both metrics).
            }

    \begin{center}
        \begin{tabular}{|c|c|c|c|c|c|}
          \hline
          {}             & \multicolumn{2}{|c|}{RaFD} & \multicolumn{2}{|c|}{CFEED} \\
          \hline
          {}             & Real or Fake        & Which's More Real   & Real or Fake  &  Which's More Real \\
          \hline
          Real        & 78.82      & -       & 72.50       & - \\
          \hline
          StarGAN~\cite{choi2018stargan}        & 31.76      & 7.06       & 14.37       & 8.75 \\
          \cline{1-5}
          GANimation~\cite{pumarola2018ganimation}     & 47.06      & 15.29       & 31.87       & 9.38  \\
          \cline{1-5}
          Ours           & \bf{74.12} & \bf{77.65}  & \bf{70.63}  & \bf{81.87}\\
          \hline
        \end{tabular}
    \end{center}
    
    \label{table:user_study}
\end{table}

\subsection{Implementation Details}
 
Our model is trained using Adam optimizer~\cite{kingma2014adam} with ${\beta}_1$ = 0.5, ${\beta}_2$ = 0.999. The detailed network architecture is provided in the supplementary materials. For a fair comparison, we train StarGAN~\cite{choi2018stargan} and GANimation~\cite{pumarola2018ganimation} using the implementations provided by the authors.
In all the experiments, StarGAN~\cite{choi2018stargan} is trained with discrete expression labels provided in the two public datasets, while GANimation~\cite{pumarola2018ganimation} 
is trained with AU intensities extracted by OpenFace toolkit~\cite{baltrusaitis2018openface}. Our network does not use any expression annotation in training.

\subsection{Quantitative Evaluation}
\label{sec:quantitative_exp}
We evaluate and compare our expression editing technique with state-of-the-art StarGAN~\cite{choi2018stargan} and GANimation~\cite{pumarola2018ganimation} quantitatively by using FID, SSIM, expression classification accuracy and user study evaluation on RaFD and CFEED.

\noindent \textbf{FID and SSIM:} Table~\ref{table:FID_SSIM} shows the evaluation results of all compared methods on the datasets RaFD and CFEED by using the FID and SSIM. As Table~\ref{table:FID_SSIM} shows, our method outperforms the state-of-the-art methods by a large margin in FID, with a 7.35 improvement on RaFD and a 5.47 improvement on CFEED. The achieved SSIMs are also higher than the state-of-the-art by 1.5$\%$ and 1.1$\%$ for the two datasets. All these results demonstrate the superior performance of our proposed LEED in synthesizing high fidelity expression images. 

\noindent \textbf{Expression Classification:} We perform quantitative evaluations with expression classification as in StarGAN~\cite{choi2018stargan} and ExprGAN~\cite{ding2018exprgan}. 
Specifically, we first train expression editing models on the training set and perform expression editing on the corresponding testing set. The edited images are then evaluated by expression classification: a higher classification accuracy means more realistic expression editing. Two classification tasks are designed: 1) train expression classifiers by using the training set images (real) and evaluate them over the edited images; 2) train classifiers by combining the real and edited images and evaluate them over the test set images. The first task evaluates whether the edited images lie in the manifold of natural expressions, and the second evaluates whether the edited images help train better classifiers.

\begin{figure}[t]
\begin{center}
\includegraphics[width=1.\linewidth]{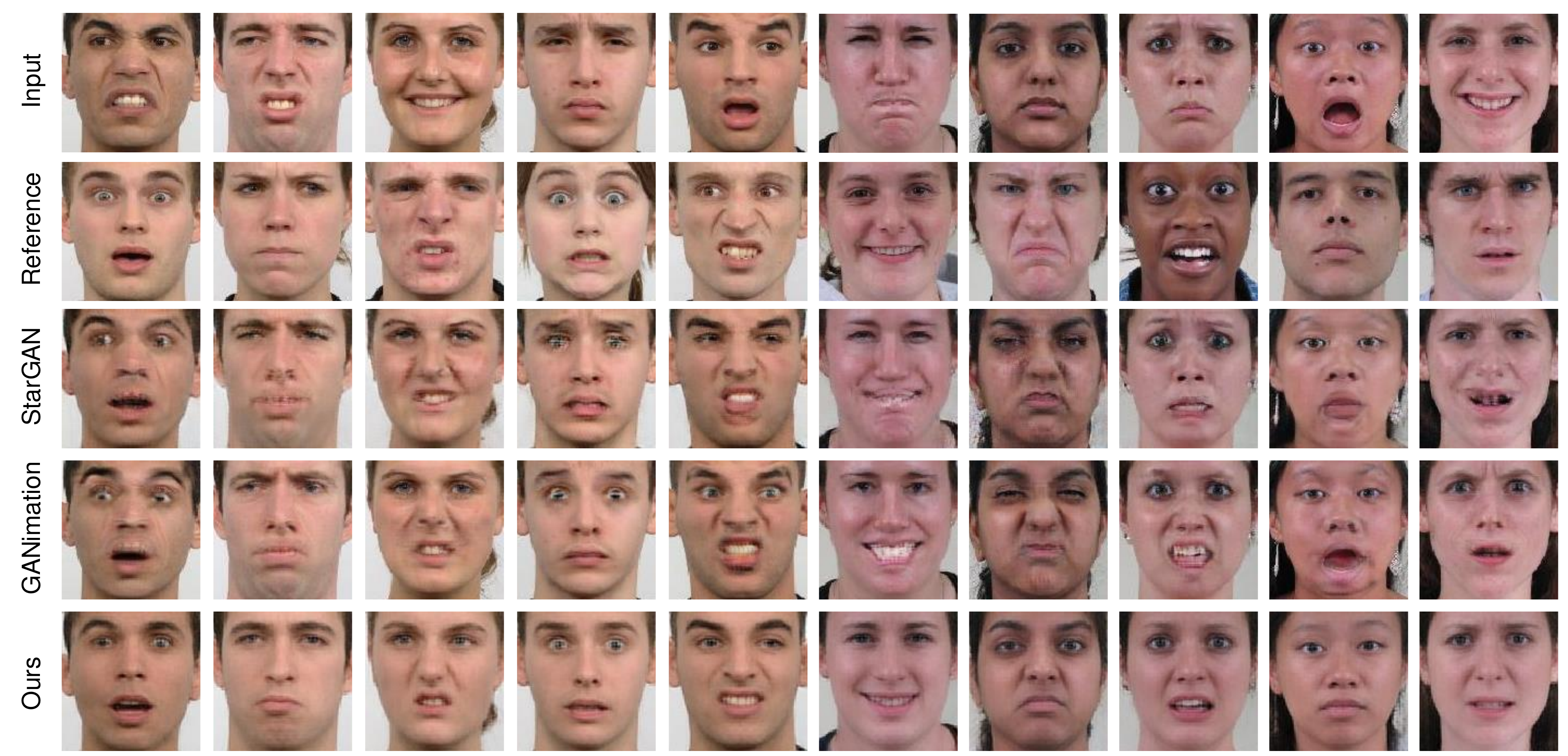}
\end{center}
\caption{
    Expression editing by LEED and state-of-the-art methods: Columns 1-5 show the editing of RaFD images, and columns 6-10 show the editing of CFEED images. Our method produces more realistic editing with better details and less artifacts. 
}

\label{fig:qualitative_evalution_combine}
\end{figure}

Table~\ref{table:cls_accuracy} shows the classification accuracy results (only seven primary expressions evaluated for CFEED). Specifically, \textbf{R} trains classifier with the original training set images and evaluates on the corresponding testing set images. \textbf{G} applies the same classifier (in \textbf{R}) to the edited images. \textbf{R + G} trains classifiers by combining the original training images and the edited ones, and evaluates on the same images in \textbf{R}. As Table~\ref{table:cls_accuracy} shows, LEED outperforms the state-of-the-art by 4.31$\%$ on RaFD and 4.89$\%$ on CFEED, respectively. Additionally, the LEED edited images help to train more accurate classifiers while incorporated in training, where the accuracy is improved by 1.04$\%$ on RaFD and 1.83$\%$ on CFEED, respectively. They also outperform StarGAN and GANimation edited images, the latter even degrade the classification probably due to the artifacts within the edited images as illustrated in Fig.~\ref{fig:qualitative_evalution_combine}. 
The two experiments demonstrate the superiority of LEED in generating more realistic expression images.

\noindent \textbf{User Studies:} We also evaluate and benchmark the LEED edited images by conducting two Amazon-Mechanical-Turk (AMT) user studies under two evaluation metrics: 1) Real or Fake: subjects are presented with a set of expression images including real ones and edited ones by LEED, GANimation, and StarGAN, and tasked to identify whether the images are real or fake; 2) Which’s More Real: subjects are presented by three randomly-ordered expression images edited by the three methods, and are tasked to identify the most real one. Table~\ref{table:user_study} shows experimental results, where LEED outperforms StarGAN and GANimation significantly under both evaluation metrics. The two user studies further demonstrate the superior perceptual fidelity of the LEED edited images.

\subsection{Qualitative Evaluation}
Fig.~\ref{fig:qualitative_evalution_combine} shows qualitative experimental results with images from RaFD (cols 1-5) and CFEED (cols 6-10). Each column shows an independent expression editing, including an input image and a reference image as well as editing by StarGAN~\cite{choi2018stargan}, GANimation~\cite{pumarola2018ganimation} and our proposed LEED.

As Fig.~\ref{fig:qualitative_evalution_combine} shows, StarGAN~\cite{choi2018stargan} and GANimation~\cite{pumarola2018ganimation} tend to generate blurs and artifacts and even corrupted facial regions (especially around eyes and mouths). LEED can instead generate more realistic facial expressions with much less blurs and artifacts, and the generated images are also clearer and sharper. In addition, LEED preserves the identity information well though it does not adopt any identity preservation loss, largely due to the identity disentanglement which encodes the identity information implicitly. 

\subsection{Ablation Study}
\begin{figure}[t]
\begin{center}
\includegraphics[width=1.\linewidth]{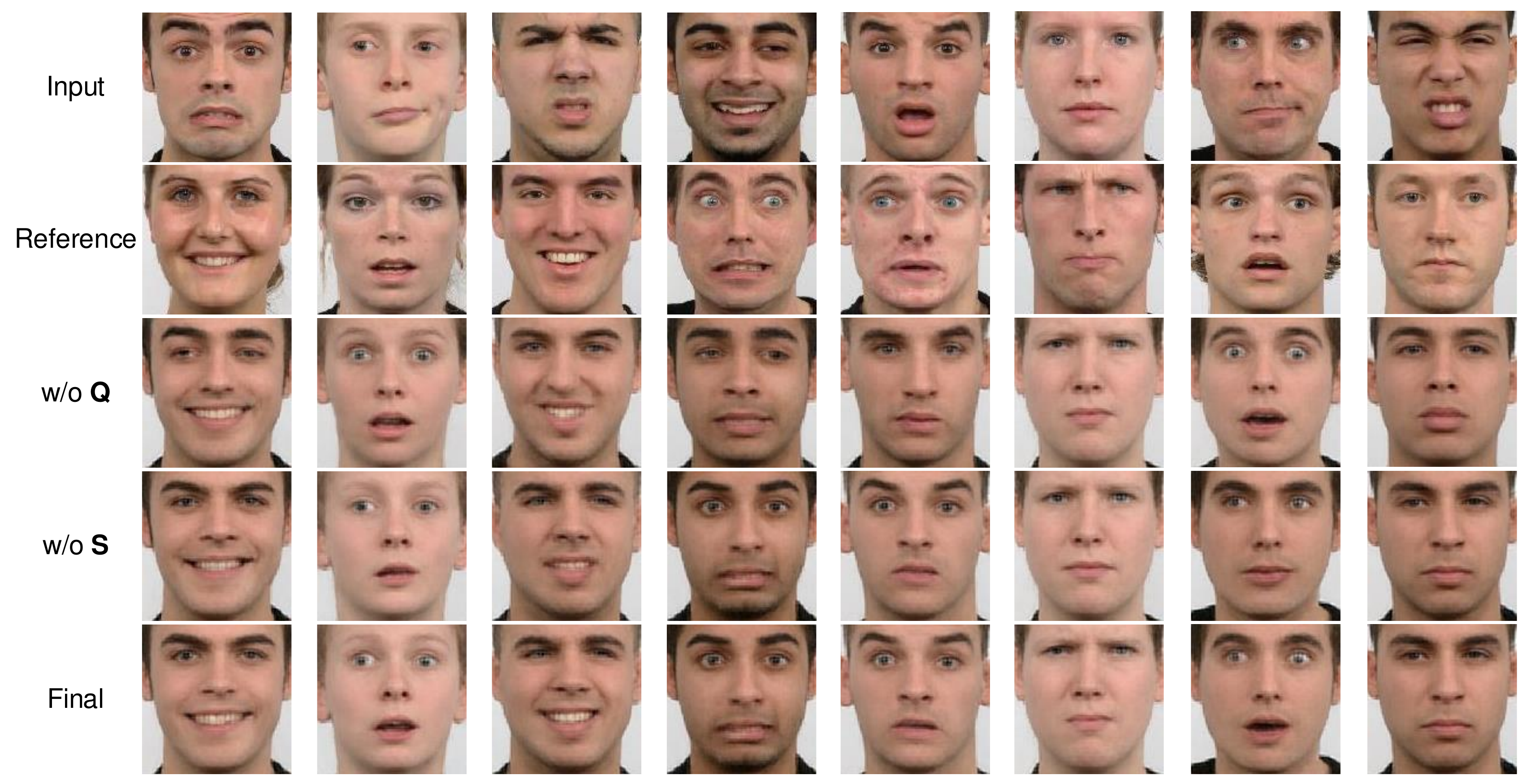}
\end{center}
\caption{
    Ablation study of LEED over RaFD: From top to bottom: input image, reference image, editing without mutual expression information loss, editing without siamese loss, final result. The graphs show the effectiveness of our designed losses.
}
\label{fig:ablation}
\end{figure}

We study the two designed losses by training three editing networks on RaFD: 1) a network without the mutual expression information loss (regularizer \bm{$Q$}) as labelled by `w/o \bm{$Q$}'; 2) a network without the siamese loss as labelled by  `w/o \bm{$S$}'; and 3) a network with both losses as labelled by `Final’ in Fig.~\ref{fig:ablation}. As Fig.~\ref{fig:ablation} shows, the mutual expression information loss guides the extractor to extract pure expression-relevant features. When it is absent, the extracted expression is degraded by expression-irrelevant information which leads to undesired editing such as eye gazing direction changes (column 1), head pose changes (columns 2, 3, 5 and 8), and identity attribute changes (missing mustache in column 4). The siamese loss enhances the expression similarity of the edited and reference images without which the expression intensity of the edited images becomes lower than that of the reference image (columns 2, 3, 6 and 7) as illustrated in Fig.~\ref{fig:ablation}.

\subsection{Discussion}
\label{discussion}

\begin{figure}[t]
\begin{center}
\includegraphics[width=1.\linewidth]{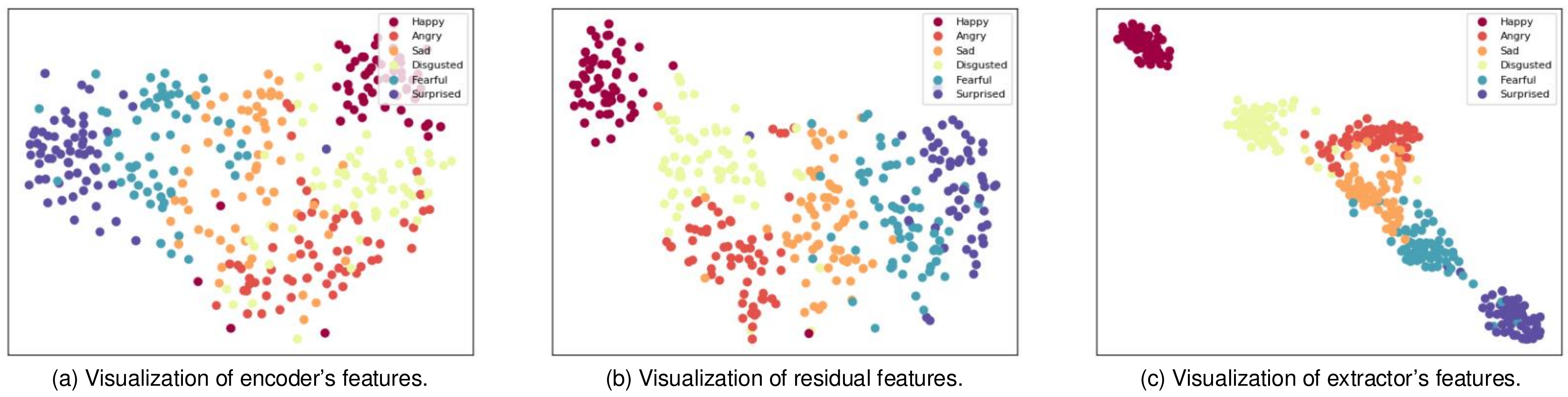}
\end{center}
\caption{   
    Expression feature Visualization with t-SNE: The \textit{Extractor} learns much more compact clusters for expression features of different classes. Best view in colors.
}
\label{fig:tsne_visualization}
\end{figure}

\noindent \textbf{Feature Visualization:} We use t-SNE~\cite{maaten2008visualizing} to show that LEED learns the right expression features via disentanglement. Besides the \textit{Extractor features}, we also show the \textit{Encoder features} (i.e. the dimension reduced representation of original image) and the \textit{Residual features} (i.e. the difference between the Encoder features of expressive and neutral faces) as illustrated in Fig.~\ref{fig:tsne_visualization} (learnt from the RaFD images). As Fig.~\ref{fig:tsne_visualization} shows, the \textit{Encoder features} and \textit{Residual features} cannot form compact expression clusters as the former learns entangled features and the latter contains expression-irrelevant features such as head-poses variations. As a comparison, the \textit{Extractor features} cluster each expression class compactly thanks to the mutual expression information loss.

\begin{figure}[t]
\centering
\begin{minipage}{.48\textwidth}
    \centering
    \includegraphics[width=.99\linewidth]{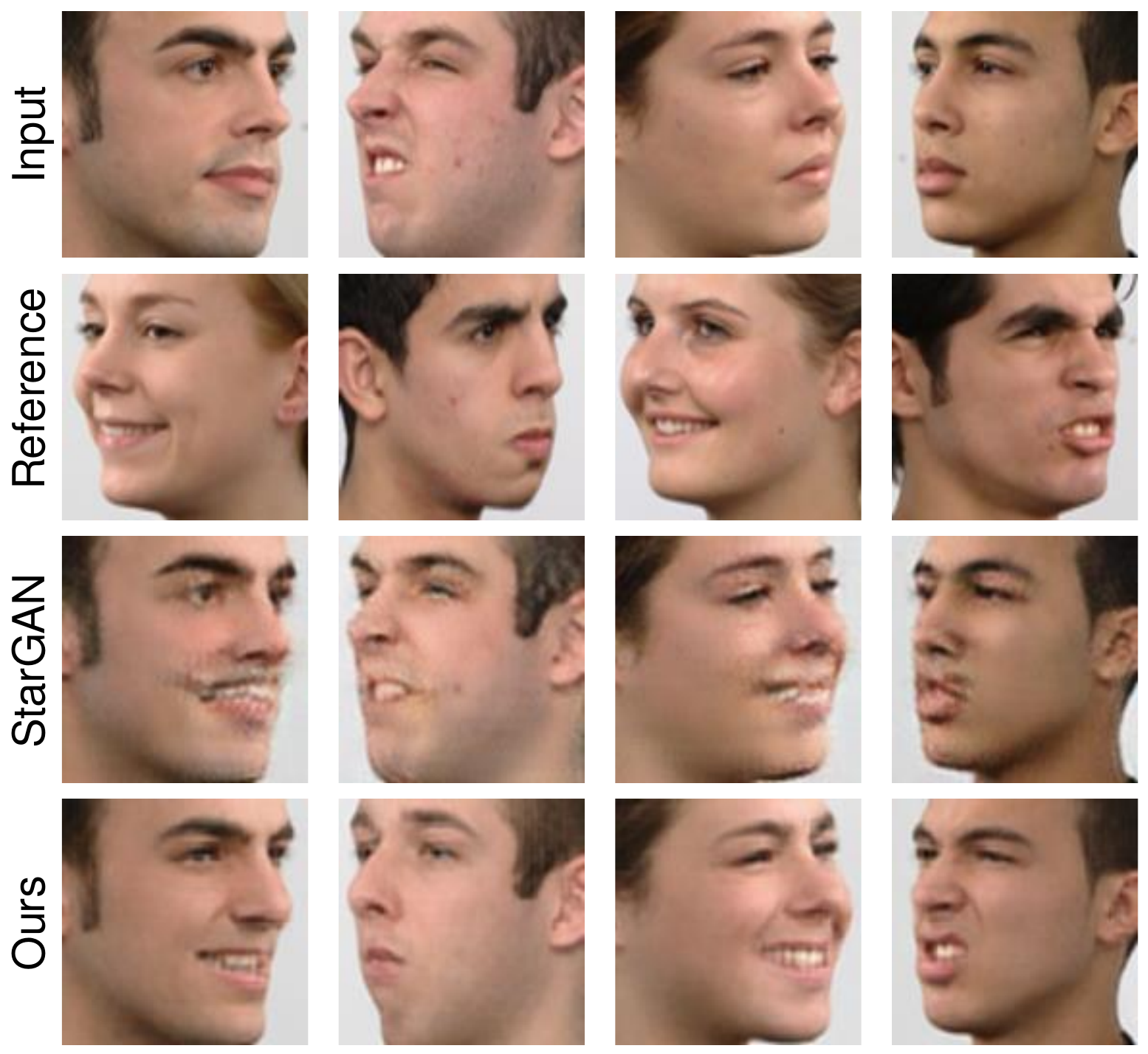}
    \caption{   
        LEED can `transfer' expressions across profile images of different poses whereas state-of-the-art StarGAN tends to produce clear artifacts.
    }
    \label{fig:rafd_profile}
\end{minipage}%
\hfill
\begin{minipage}{.48\textwidth}
    \centering
    \includegraphics[width=.99\linewidth]{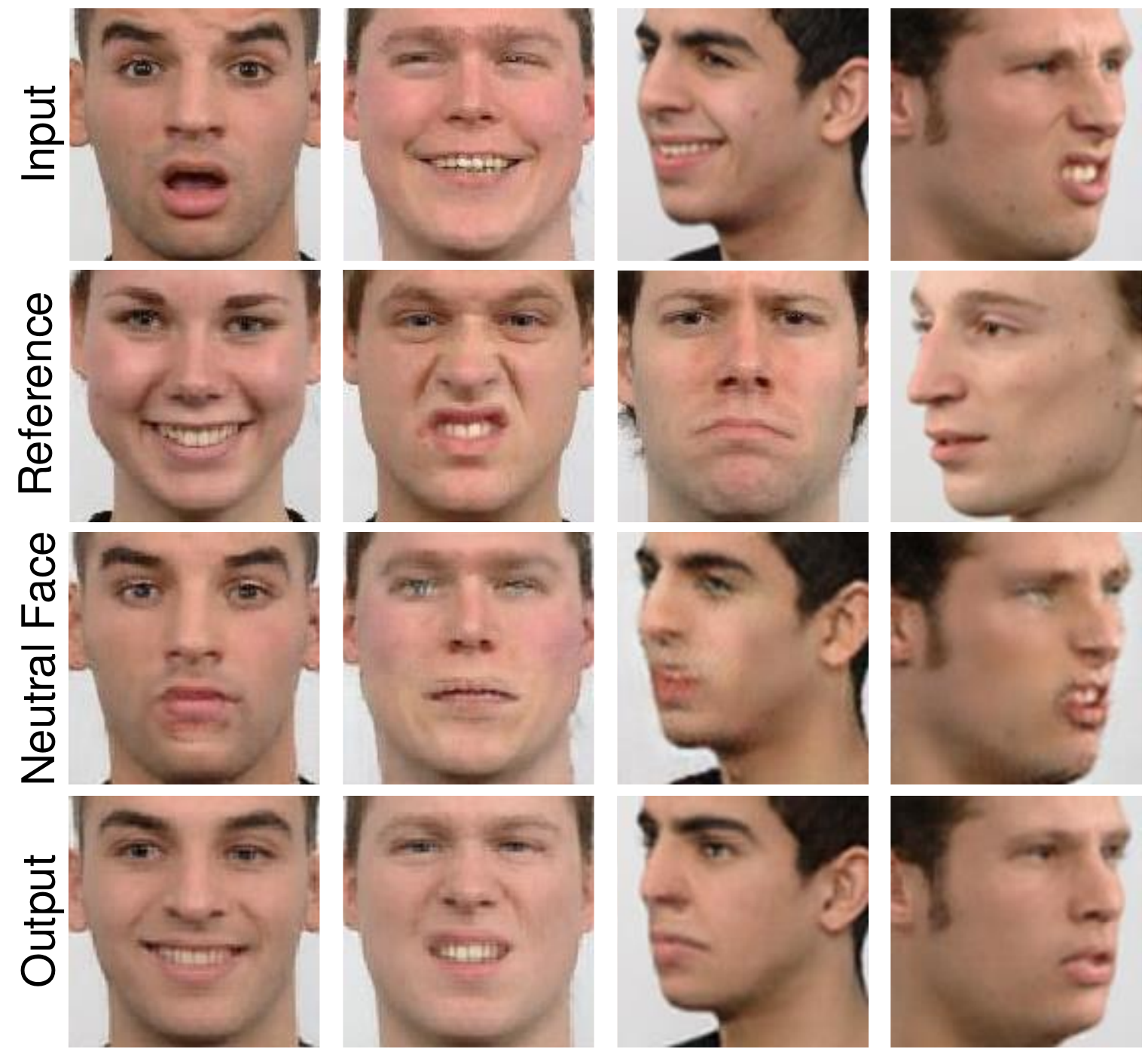}
    \caption{
        Though the GAN-generated neutral faces may not be perfect, LEED still generate sharp and clear expression thanks to our adopted adversarial loss. 
    }
    \label{fig:bad_neutral}
\end{minipage}
\end{figure}

\noindent \textbf{Expression Editing on Profile Images:} LEED is capable of `transferring’ expression across profile images of different poses. As illustrated in Fig.~\ref{fig:rafd_profile}, LEED produces realistic expression editing with good detail preservation whereas StarGAN introduces lots of artifacts 
(GANimation does not work as OpenFace cannot extract AUs accurately from profile faces). The capability of handling profile images is largely attributed to the mutual expression information loss that helps extract expression related features 
in the reference image. Note AF-VAE~\cite{qian2019make} can also work with non-frontal profile images but it can only transfer expressions across facial images of the same pose.

\noindent \textbf{Robustness to Imperfect Neutral Expression Images:} LEED uses a pre-train GAN to generate neutral expression images for the disentanglement but the generated neural face may not be perfect as illustrated in Fig.~\ref{fig:bad_neutral} (row 3). LEED is tolerant to such imperfection as shown in Fig.~\ref{fig:bad_neutral} (row 4), largely because of the adversarial loss that is included into the interpolated latent code.
However, the imperfect neutral face of reference image may contain residual of the original expression and lead to lower expression intensity in the output. This issue could be mitigated by adopting a stronger expression normalization model.

\begin{figure}[t]
\begin{center}
\includegraphics[width=1.\linewidth]{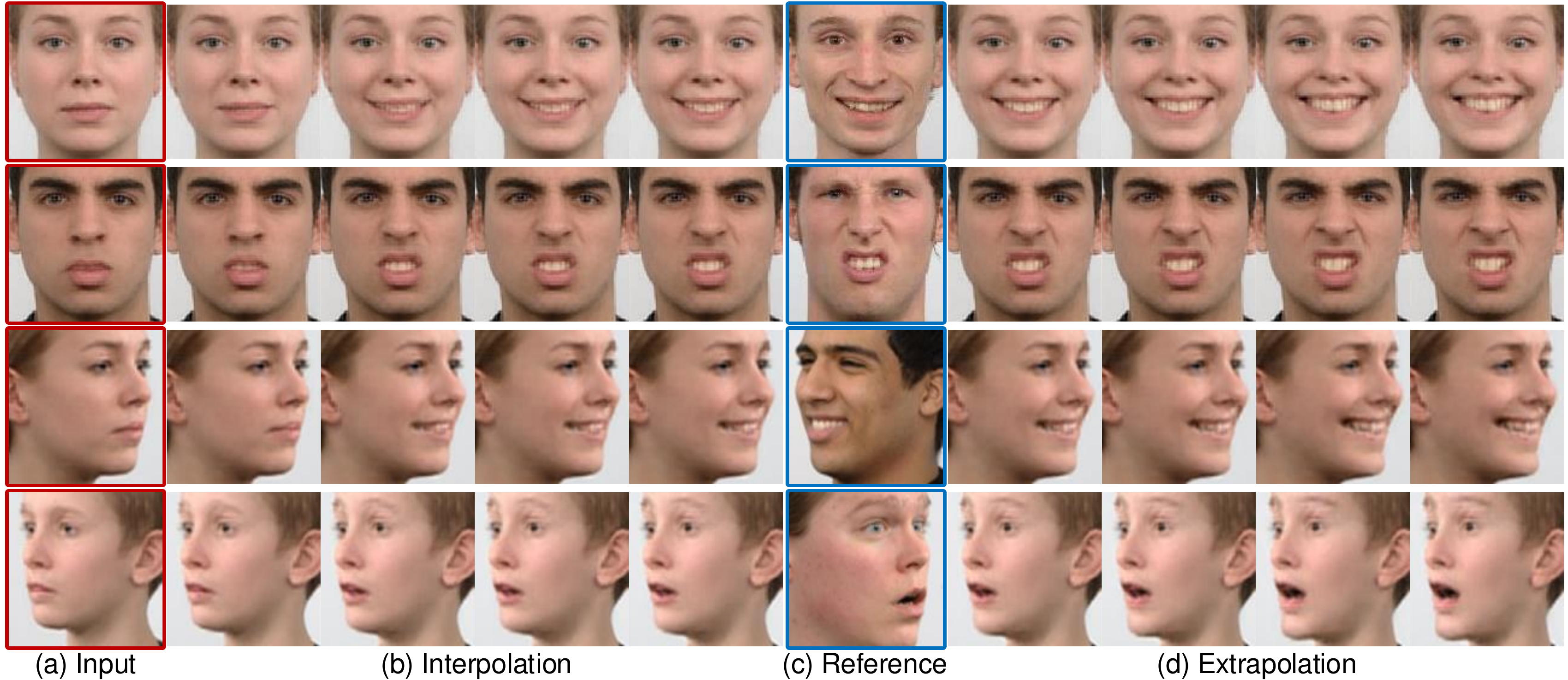}
\end{center}   
\caption{
    Expression editing via interpolation/extrapolation: Given input images in (a) and reference images in (c), LEED can edit expressions by either interpolation ($\alpha <$ 1) or extrapolation ($\alpha >$ 1) as shown in (b) and (d).
}
\label{fig:sequence}
\end{figure}

\noindent \textbf{Continuous Editing:} Our method can generate continuous expression sequences by changing the interpolated factor $\alpha$ (Sec.~\ref{Extractor_Interpolator}) as shown in Fig.~\ref{fig:sequence}. 
Besides interpolation, we show that the extrapolation can generate extreme expressions. This shows our method could uncover the structure of natural expression manifolds.

\begin{figure}[t]
\begin{center}
\includegraphics[width=1.\linewidth]{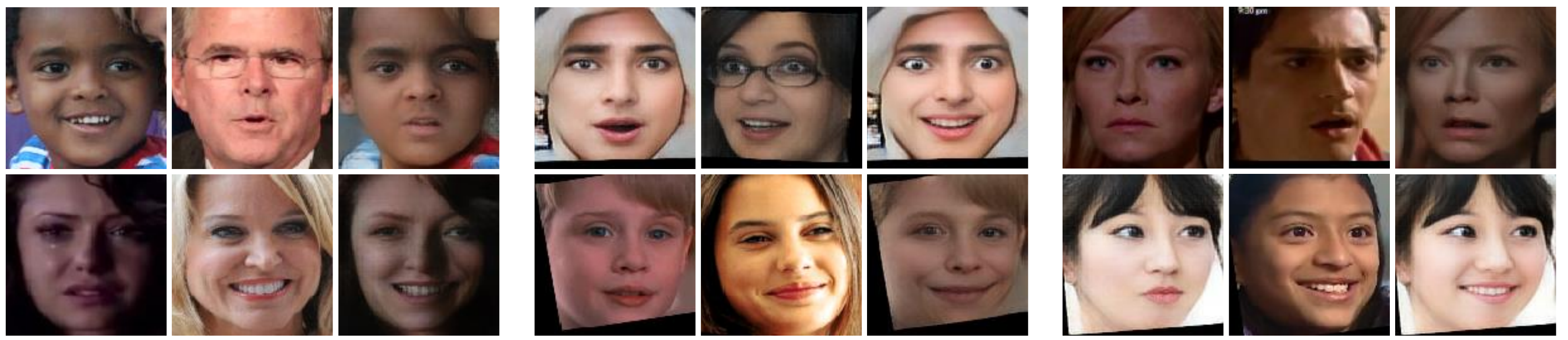}
\end{center}
\caption{
    Facial expression editing by LEED on wild images: In each triplet, the first column is input facial image, the second column is the image with desired expression and the last column is the synthesized result.
}
\label{fig:exp_wild}
\end{figure}

\noindent \textbf{Facial Expression Editing on Wild Images:} FEE for wild images is much more challenging as the images have more variations in complex background, uneven lighting, etc. LEED can adapt to handle wild images well as illustrated in Fig.~\ref{fig:exp_wild}, where the model is trained on expressive images sampled from AffectNet~\cite{mollahosseini2017affectnet}.
As Fig.~\ref{fig:exp_wild} shows, LEED can transform the expressions successfully while maintaining the expression-unrelated information unchanged.

\section{Conclusion}
We propose a novel label-free expression editing via disentanglement (LEED) framework for realistic expression editing of both frontal and profile facial images without any expression annotation. Our method disentangles the identity and expression of facial images and edits expressions by fusing the identity of the input image and the expression of the reference image. Extensive experiments over two public datasets show that LEED achieves superior expression editing as compared with the state-of-the-art techniques. 
We expect that LEED will inspire new insights and attract more interests for better FEE in the near future.

\section{Acknowledgement}
This work is supported by Data Science \& Artificial Intelligence Research Centre, NTU Singapore.

\clearpage
%
%
\bibliographystyle{splncs04}
\bibliography{egbib}

\clearpage

\section{Network Architecture}
Our network has five major components: an extractor \bm{$\mathcal{X}$} for extracting expression attribute from the reference image; an interpolator \bm{$\mathcal{I}$} for
fusing the extracted expression attribute and the identity attribute of the input image; an encoder \bm{$E$} for mapping the facial images into a compact expression and identity embedded space; a decoder \bm{$D$} for projecting the interpolated code to image space and a pre-trained GAN for synthesizing the neutral faces. Besides, a discriminator \bm{$\mathcal{D}$} is designed for distinguishing the real/interpolated codes, a regularizer \bm{$Q$} and siamese network \bm{$S$} for optimal expression disentanglement and consistent synthesis, respectively. The detailed architectures are shown in Tables 1-6~\footnote{We pretrain StarGAN~\cite{choi2018stargan} on the corresponding dataset and use it for synthesizing neutral faces, with official implementation at \url{https://github.com/yunjey/stargan}.}.

\section{Training Details}
We adopt Adam optimizer with ${\beta}_1$ = 0.5, ${\beta}_2$ = 0.999 for optimization. 
We set $\lambda_{Q}$, $\lambda_{recon}$, $\lambda_{S}$ and $\lambda_{gp}$ to be 0.01, 10, 1000 and 100 to balance the magnitude of different losses.
The batchsize is set to 24. The total number of epochs is set to 100.
The initial learning rate is set to 1e-4 for the first 50 epochs, then linearly decay to 0 over another 50 epochs.
The training process takes 7 hours on RaFD~\cite{langner2010presentation} and 13 hours on CFEED~\cite{du2014compound} on a single Tesla V100 GPU, respectively.



\section{More Results}
We also present more results generated by LEED in the following pages.


\clearpage

\begin{figure}[t]
\begin{center}
\includegraphics[width=1\linewidth]{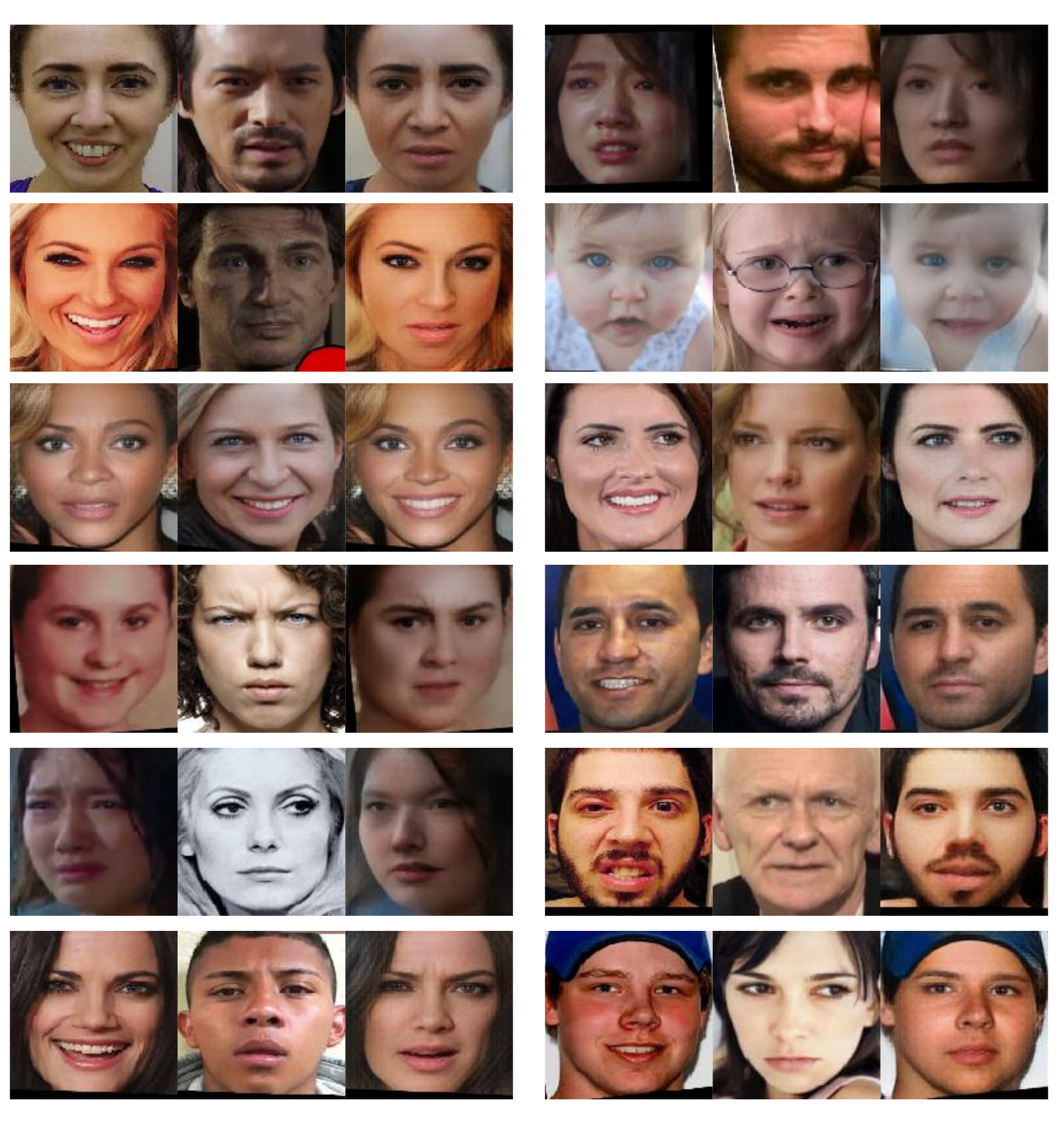}
\end{center}
\caption{   
    Additional expression editing results on wild images. In each triplet, the first column is input facial image, the second column is the image with desired expression and the last column is the synthesized result.
}
\label{fig:wild_qualitative}
\end{figure}

\begin{figure}[t]
\begin{center}
\includegraphics[width=1\linewidth]{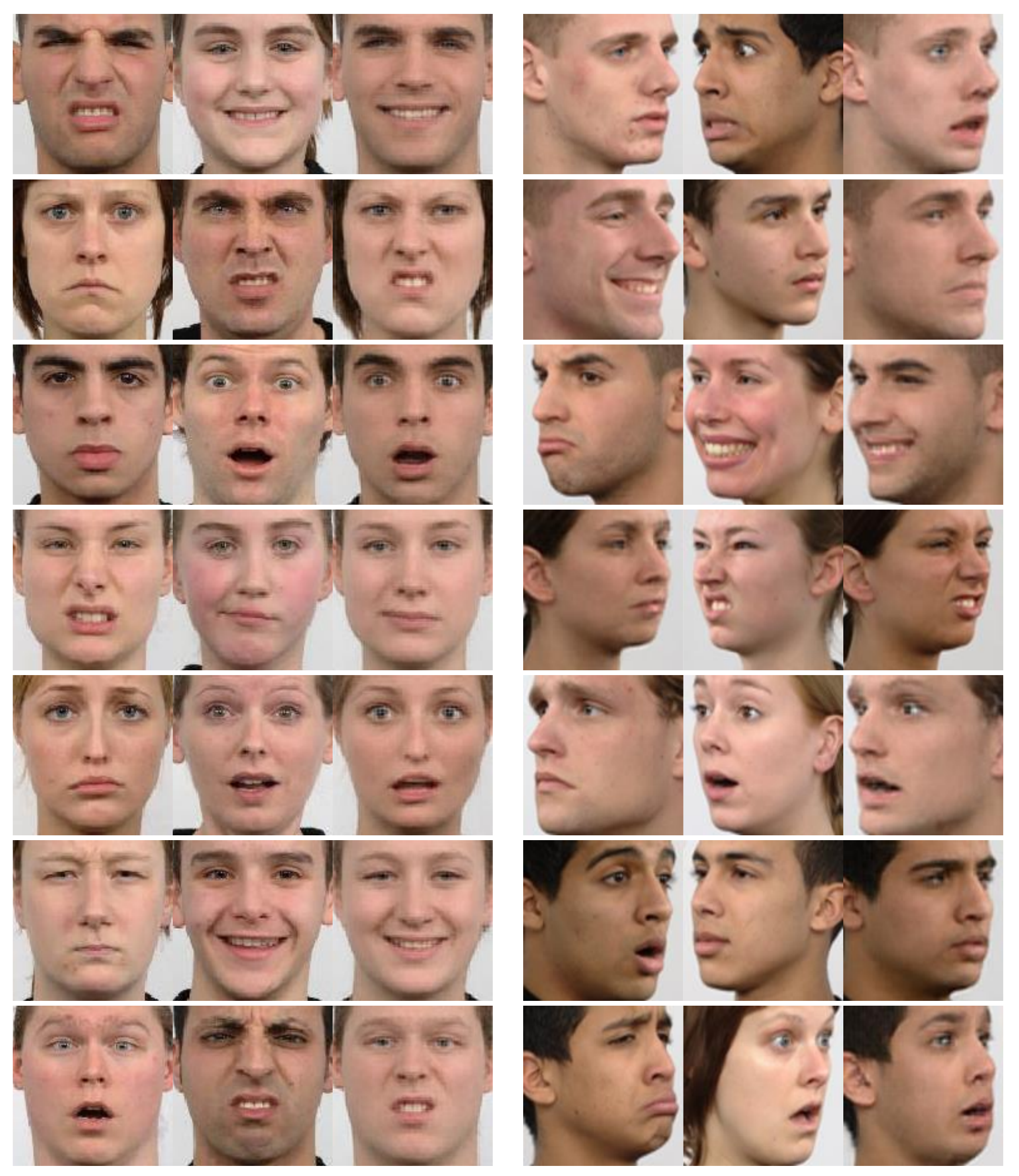}
\end{center}
\caption{   
    Additional expression editing results on RaFD~\cite{langner2010presentation}. In each triplet, the first column is input facial image, the second column is the image with desired expression and the last column is the synthesized result.
}
\label{fig:rafd_qualitative}
\end{figure}

\begin{figure}[t]
\begin{center}
\includegraphics[width=1\linewidth]{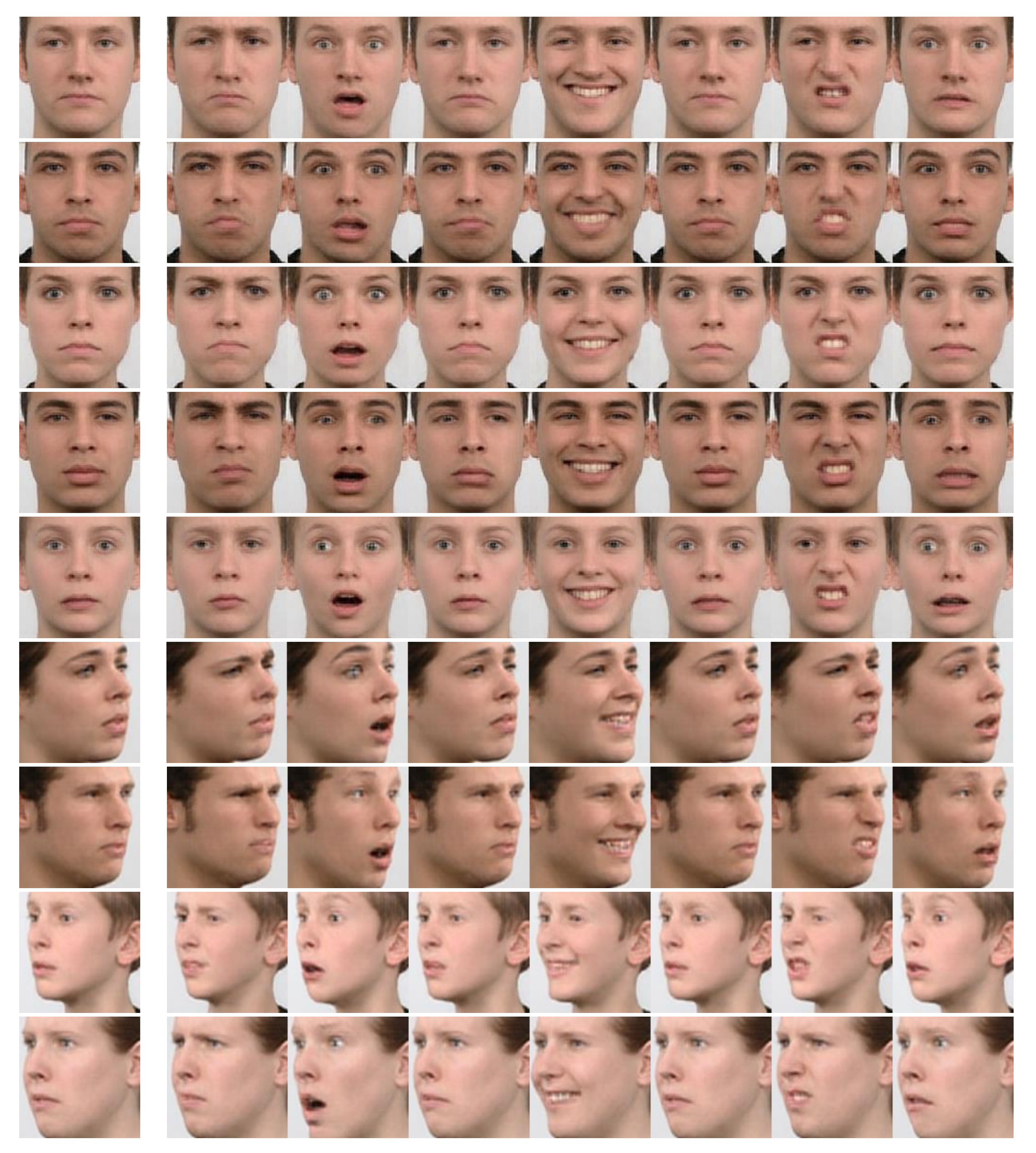}
\end{center}
\caption{   
    Additional expression editing results on RaFD~\cite{langner2010presentation}. For each row, the left most one is the input facial image, and the rest gives the synthesized expressions (Angry, Surprised, Sad, Happy, Neutral, Disgusted, Fearful).
}
\label{fig:rafd_different_expr}
\end{figure}

\begin{figure}[t]
\begin{center}
\includegraphics[width=1\linewidth]{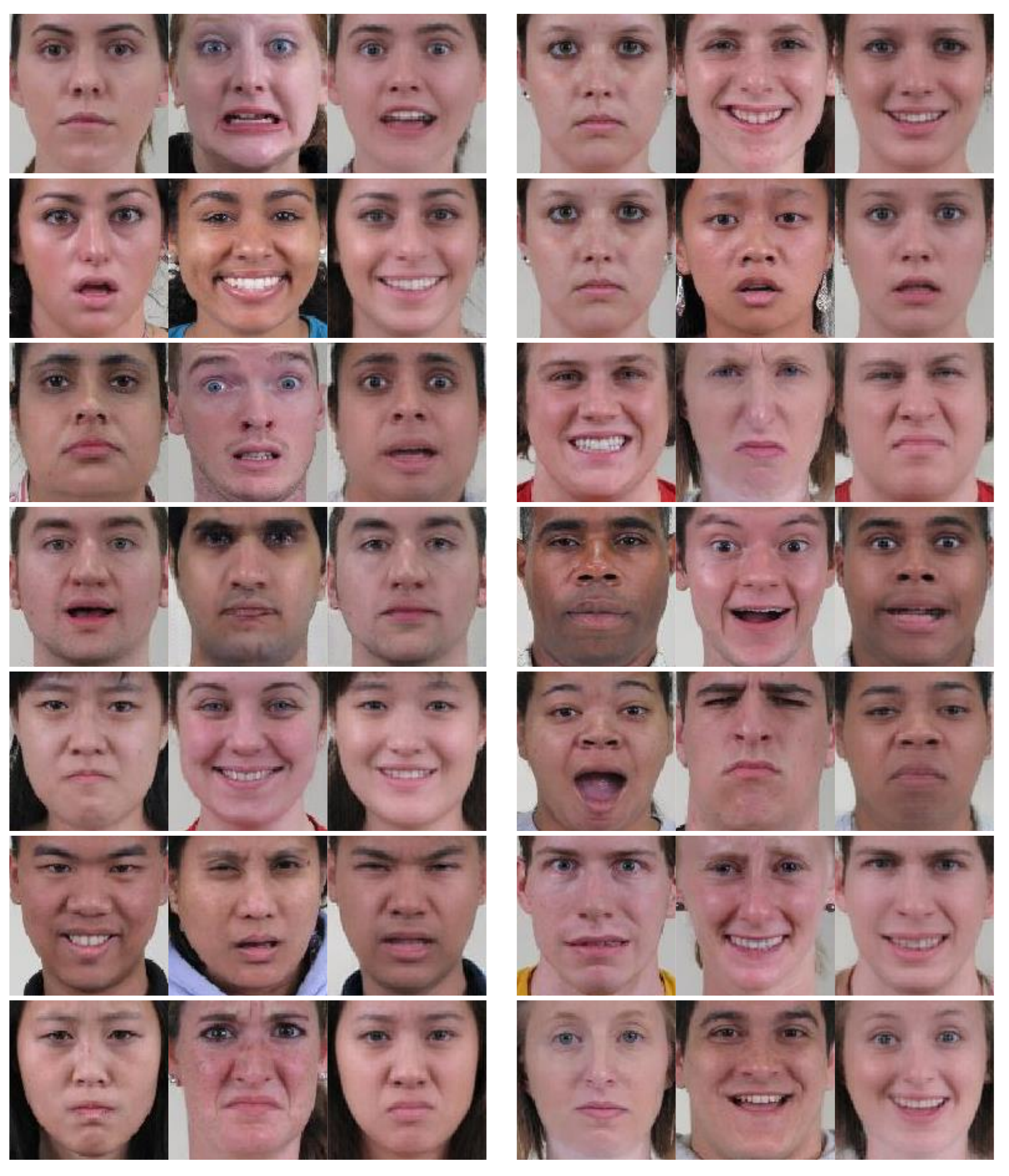}
\end{center}
\caption{   
    Additional expression editing results on CFEED~\cite{du2014compound}. In each triplet, the first column is input facial image, the second column is the image with desired expression and the last column is the synthesized result.
}
\label{fig:cfeed_qualitative}
\end{figure}

\begin{figure}[t]
\begin{center}
\includegraphics[width=1\linewidth]{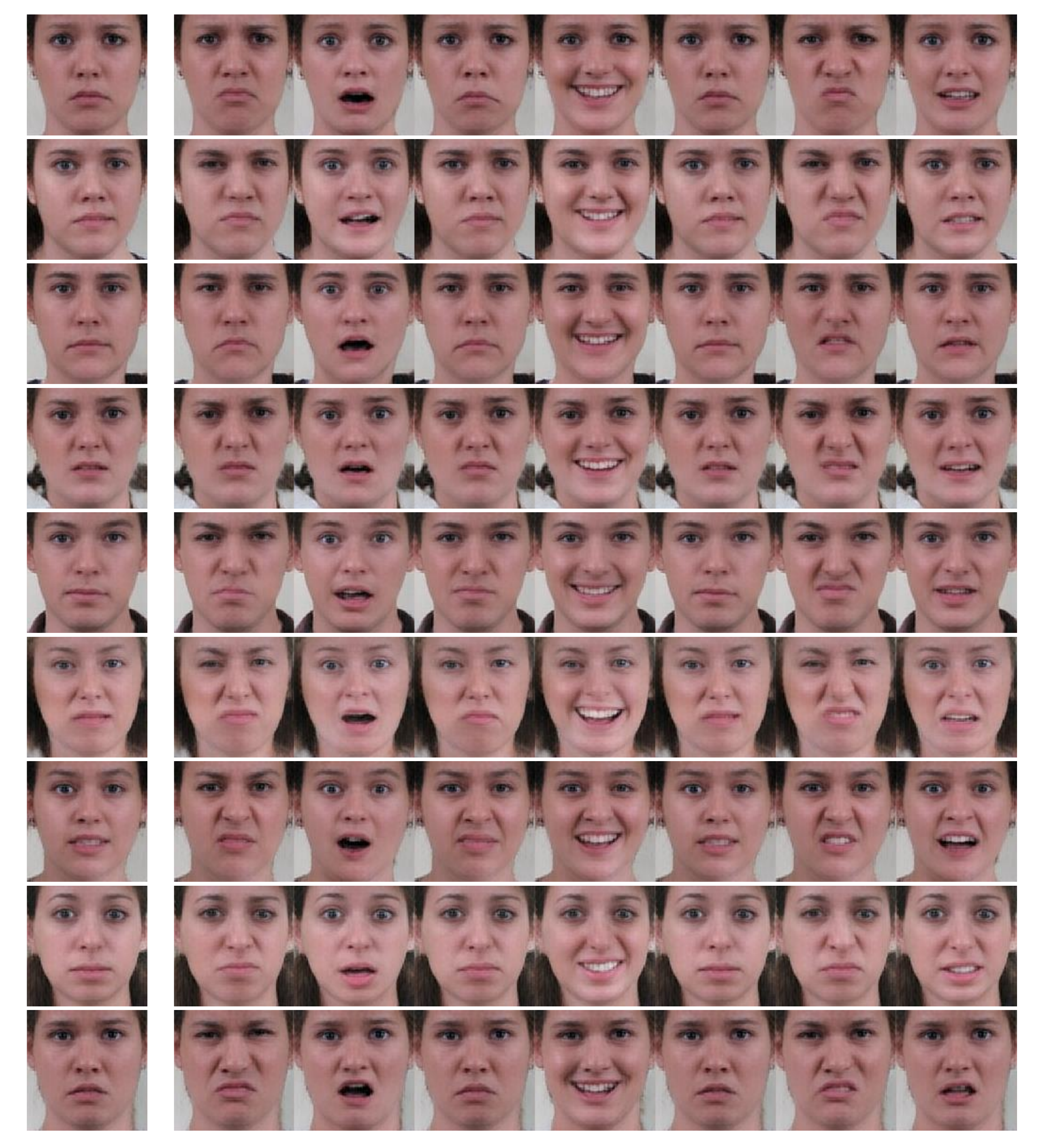}
\end{center}
\caption{   
    Additional expression editing results on CFEED~\cite{du2014compound}. For each row, the left most one is the input facial image, and the rest gives the synthesized expressions (Angry, Surprised, Sad, Happy, Neutral, Disgusted, Fearful).
}
\label{fig:cfeed_different_expr}
\end{figure}

\clearpage
\newpage

\begin{table}[t]
    \caption{
            Architecture of extractor \bm{$\mathcal{X}$} and interpolator \bm{$\mathcal{I}$}. \bm{$\mathcal{X}$} and \bm{$\mathcal{I}$} share the same architecture.
            }

    \begin{center}
        \begin{tabular}{cccccccc}
          \hline
          Layer Type & Output Size & Channel & Kernel & Stride & Padding & Normalization & Activation \\
          \hline
          Conv2d & 512 $\times$ 8 $\times$ 8 & 512 & 3 & 1 & 1 & - & LeakyReLU \\
          Conv2d & 512 $\times$ 8 $\times$ 8 & 512 & 3 & 1 & 1 & - & LeakyReLU \\
          Conv2d & 512 $\times$ 8 $\times$ 8 & 512 & 3 & 1 & 1 & - & - \\
          \hline
        \end{tabular}
    \end{center}
    
    \label{table:architecture_extractor_interpolator}
\end{table}

\begin{table}[t]
    \caption{
            Architecture of discriminator \bm{$\mathcal{D}$}. IN stands for instance normalization.
            }

    \begin{center}
        \begin{tabular}{cccccccc}
          \hline
          Layer Type & Output Size & Channel & Kernel & Stride & Padding & Normalization & Activation \\
          \hline
          Conv2d & 256 $\times$ 8 $\times$ 8 & 256 & 1 & 1 & 0 & IN & LeakyReLU \\
          Conv2d & 512 $\times$ 4 $\times$ 4 & 512 & 4 & 2 & 1 & IN & LeakyReLU \\
          Conv2d & 1024 $\times$ 2 $\times$ 2 & 1024 & 4 & 2 & 1 & IN & LeakyReLU \\
          Conv2d & 1024 $\times$ 1 $\times$ 1 & 1024 & 2 & 2 & 0 & - & - \\
          \hline
        \end{tabular}
    \end{center}
    
    \label{table:architecture_discriminator}
\end{table}

\begin{table}[t]
    \caption{
            Architecture of encoder \bm{$E$}.
            }

    \begin{center}
        \begin{tabular}{cccccccc}
          \hline
          Layer Type & Output Size & Channel & Kernel & Stride & Padding & Normalization & Activation \\
          \hline
          Conv2d & 64 $\times$ 128 $\times$ 128 & 64 & 3 & 1 & 1 & - & ReLU \\
          Conv2d & 64 $\times$ 128 $\times$ 128 & 64 & 3 & 1 & 1 & - & ReLU \\ 
          MaxPool2d & 64 $\times$ 64 $\times$ 64 & - & 2 & 2 & 0 & - & -  \\
          Conv2d & 128 $\times$ 64 $\times$ 64 & 128 & 3 & 1 & 1 & - & ReLU \\
          Conv2d & 128 $\times$ 64 $\times$ 64 & 128 & 3 & 1 & 1 & - & ReLU \\ 
          MaxPool2d & 128 $\times$ 32 $\times$ 32 & - & 2 & 2 & 0 & - & -  \\
          Conv2d & 256 $\times$ 32 $\times$ 32 & 256 & 3 & 1 & 1 & - & ReLU \\
          Conv2d & 256 $\times$ 32 $\times$ 32 & 256 & 3 & 1 & 1 & - & ReLU \\ 
          Conv2d & 256 $\times$ 32 $\times$ 32 & 256 & 3 & 1 & 1 & - & ReLU \\
          Conv2d & 256 $\times$ 32 $\times$ 32 & 256 & 3 & 1 & 1 & - & ReLU \\ 
          MaxPool2d & 256 $\times$ 16 $\times$ 16 & - & 2 & 2 & 0 & - & -  \\
          Conv2d & 512 $\times$ 16 $\times$ 16 & 512 & 3 & 1 & 1 & - & ReLU \\
          Conv2d & 512 $\times$ 16 $\times$ 16 & 512 & 3 & 1 & 1 & - & ReLU \\ 
          Conv2d & 512 $\times$ 16 $\times$ 16 & 512 & 3 & 1 & 1 & - & ReLU \\
          Conv2d & 512 $\times$ 16 $\times$ 16 & 512 & 3 & 1 & 1 & - & ReLU \\ 
          MaxPool2d & 512 $\times$ 8 $\times$ 8 & - & 2 & 2 & 0 & - & -  \\
          Conv2d & 512 $\times$ 8 $\times$ 8 & 512 & 3 & 1 & 1 & - & - \\ 
          \hline
        \end{tabular}
    \end{center}
    
    \label{table:architecture_encoder}
\end{table}

\begin{table}[t]
    \caption{
            Architecture of decoder \bm{$D$}. BN stands for batch normalization.
            }

    \begin{center}
        \begin{tabular}{cccccccc}
          \hline
          Layer Type & Output Size & Channel & Kernel & Stride & Padding & Normalization & Activation \\
          \hline
          Conv2d & 512 $\times$ 8 $\times$ 8 & 512 & 3 & 1 & 1 & BN & ReLU \\
          Upsample & 512 $\times$ 16 $\times$ 16 & - & - & - & - & - & - \\ 
          Conv2d & 256 $\times$ 16 $\times$ 16 & 256 & 3 & 1 & 1 & BN & ReLU  \\
          Conv2d & 256 $\times$ 16 $\times$ 16 & 256 & 3 & 1 & 1 & BN & ReLU  \\
          Conv2d & 256 $\times$ 16 $\times$ 16 & 256 & 3 & 1 & 1 & BN & ReLU  \\
          Conv2d & 256 $\times$ 16 $\times$ 16 & 256 & 3 & 1 & 1 & BN & ReLU  \\
          Upsample & 256 $\times$ 32 $\times$ 32 & - & - & - & - & - & - \\ 
          Conv2d & 128 $\times$ 32 $\times$ 32 & 128 & 3 & 1 & 1 & BN & ReLU  \\
          Conv2d & 128 $\times$ 32 $\times$ 32 & 128 & 3 & 1 & 1 & BN & ReLU  \\
          Conv2d & 128 $\times$ 32 $\times$ 32 & 128 & 3 & 1 & 1 & BN & ReLU  \\
          Conv2d & 128 $\times$ 32 $\times$ 32 & 128 & 3 & 1 & 1 & BN & ReLU  \\
          Upsample & 128 $\times$ 64 $\times$ 64 & - & - & - & - & - & - \\ 
          Conv2d & 64 $\times$ 64 $\times$ 64 & 64 & 3 & 1 & 1 & BN & ReLU  \\
          Conv2d & 64 $\times$ 64 $\times$ 64 & 64 & 3 & 1 & 1 & BN & ReLU  \\
          Upsample & 64 $\times$ 128 $\times$ 128 & - & - & - & - & - & - \\
          Conv2d & 64 $\times$ 128 $\times$ 128 & 64 & 3 & 1 & 1 & BN & ReLU  \\
          Conv2d & 3 $\times$ 128 $\times$ 128 & 3 & 3 & 1 & 1 & - & -  \\
          \hline
        \end{tabular}
    \end{center}
    
    \label{table:architecture_decoder}
\end{table}

\begin{table}[t]
    \caption{
            Architecture of regularizer \bm{$Q$}. BN stands for batch normalization.
            }

    \begin{center}
        \begin{tabular}{cccccccc}
          \hline
          Layer Type & Output Size & Channel & Kernel & Stride & Padding & Normalization & Activation \\
          \hline
          Conv2d & 512 $\times$ 1 $\times$ 1 & 512 & 8 & 1 & 0 & BN & LeakyReLU \\
          FC & 128  & 128 & - & - & - & BN & LeakyReLU \\
          FC & 16  & 16 & - & - & - & - & - \\
          \hline
        \end{tabular}
    \end{center}
    
    \label{table:architecture_Q}
\end{table}

\begin{table}[t]
    \caption{
            Architecture of siamese network \bm{$S$}. IN stands for instance normalization.
            }

    \begin{center}
        \begin{tabular}{cccccccc}
          \hline
          Layer Type & Output Size & Channel & Kernel & Stride & Padding & Normalization & Activation \\
          \hline
          Conv2d & 64 $\times$ 64 $\times$ 64 & 64 & 4 & 2 & 1 & IN & LeakyReLU \\
          Conv2d & 128 $\times$ 32 $\times$ 32 & 128 & 4 & 2 & 1 & IN & LeakyReLU \\
          Conv2d & 256 $\times$ 16 $\times$ 16 & 256 & 4 & 2 & 1 & IN & LeakyReLU \\
          Conv2d & 512 $\times$ 8 $\times$ 8 & 512 & 4 & 2 & 1 & IN & LeakyReLU \\
          Conv2d & 1024 $\times$ 4 $\times$ 4 & 1024 & 4 & 2 & 1 & IN & LeakyReLU \\
          Conv2d & 2048 $\times$ 2 $\times$ 2 & 1024 & 4 & 2 & 1 & IN & LeakyReLU \\
          FC & 1024 & 1024 & - & - & - & - & - \\
          \hline
        \end{tabular}
    \end{center}
    
    \label{table:architecture_S}
\end{table}

\end{document}